\documentclass{llncs}

\usepackage{llncs-space-min}

\usepackage{graphicx}
\usepackage{wrapfig}
\usepackage{amsmath}
\usepackage{amssymb}
\usepackage{mathtools}
\usepackage{xcolor,colortbl}
\usepackage{adjustbox}
\usepackage{multirow}
\usepackage{booktabs}
\usepackage{siunitx}
\usepackage{listings}
\usepackage{url}

\usepackage{pxfonts}
\usepackage{anyfontsize}

\definecolor{LightGray}{gray}{0.85}
\definecolor{DarkGray}{gray}{0.65}

\newcolumntype{a}{>{\columncolor{LightGray}}c}

\begin{document}

\title{Global and local evaluation of link prediction tasks with neural embeddings}
%
\titlerunning{Global and local evaluation of link prediction tasks with neural embeddings}  
%
\author{Asan Agibetov\inst{1} 
 \and Matthias Samwald\inst{1}}
\authorrunning{Asan Agibetov et al.} 
%
%
\institute{$^1$Section for Artificial Intelligence and Decision Support; Center for Medical Statistics, Informatics, and Intelligent Systems; Medical University of Vienna, Austria\\
\email{asan.agibetov@meduniwien.ac.at}
}

\maketitle              

\begin{abstract} 

    We focus our attention on the link prediction problem for knowledge graphs, which is treated herein as a binary classification task on neural embeddings of the
    entities. By comparing, combining and extending different methodologies for link prediction on graph-based data coming from different domains, we formalize a unified
    methodology for the quality evaluation benchmark of neural embeddings for knowledge graphs. This benchmark is then used to empirically investigate the potential of
    training neural embeddings globally for the entire graph, as opposed to the usual way of training embeddings locally for a specific relation. This new way of testing
    the quality of the embeddings evaluates the performance of binary classifiers for scalable link prediction with limited data. Our evaluation pipeline is made open
    source, and with this we aim to draw more attention of the community towards an important issue of transparency and reproducibility of the neural
    embeddings evaluations.

\end{abstract}
\section{Introduction}
\subsubsection{Link prediction} \label{sec:intro-lp}

There are two major ways of measuring the quality of (neural) embeddings of entities in a Knowledge Graph for link prediction tasks, inspired by two different
fields: information retrieval~\cite{bordes_2013,nickel_2016,kadlec_2017,wu_2017,nickel_2017} and graph-based data
mining~\cite{perozzi_2014,grover_2016,garciagasulla_2016,chamberlain_2017,alshahrani_2017,agibetov_2018}. Information retrieval inspired approaches seem to favor the mean
rank measurement and its variants (mean average precision, top $k$ results, mean reciprocal rank), and graph-based data mining approaches recur to the standard evaluation
measurements of classifiers based on false positive rate to true positive rate curves (e.g., ROC AUC, F-measure). While both of the techniques measure the quality of the
embeddings, they however may or may not be suitable for particular link prediction tasks. Consider the popular mean rank (and its variants) that measures the ability
of a retrieval system to score a \emph{true} result on top of all other possible candidates.
Applied to the knowledge graph completion task~\cite{bordes_2013}, typically, one would consider all links, i.e., assertions of relations of type $(e_i, r_k, e_j) \in
KG$, as \emph{true} and all other possible assertions that are not in the Knowledge Graph as \emph{negative} (i.e., $(\bar{e_i}, r_i, \bar{e_j}) \not \in KG$).  If we use
mean rank as our metric to measure the performance of a classifier for link prediction we might have the following problems. 

\begin{wrapfigure}{l}{0.5\textwidth}
    \begin{center}
        \includegraphics[width=.5\textwidth]{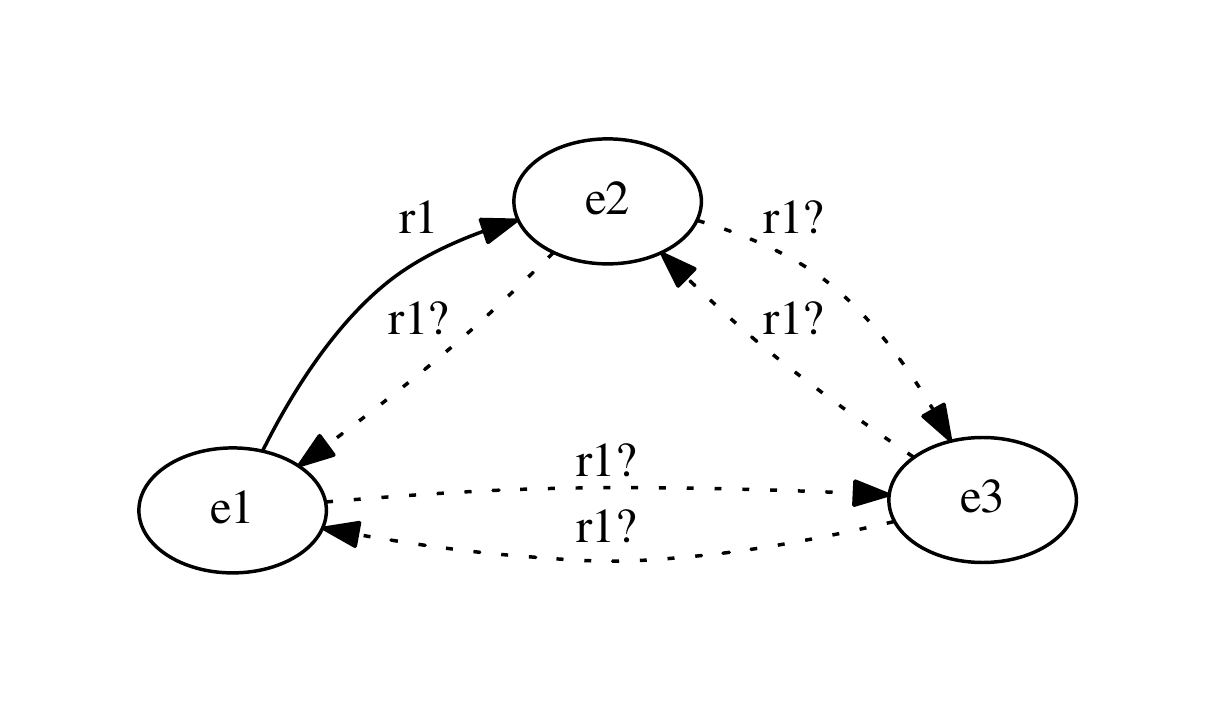}
    \end{center}
    \caption{\label{fig:simple-lp-task} Simple KG and a link prediction task. Bold links known to exist, dashed links are unknown.}
\end{wrapfigure}

Consider a simple KG, consisting of three
entities $e_1, e_2, e_3$ and one relation $r_1$, and assume that we only know about the existence of the link $(e_1, r_1, e_2)$, as depicted in
Figure~\ref{fig:simple-lp-task}. Our only true link (example for a prediction system) is $t_1 = (e_1, r_1, e_2)$ and our negative links are $n_1 = (e_1, r_1, e_3), n_2 =
(e_3, r_1, e_1), n_3 = (e_2, r_1, e_3), n_4 = (e_3, r_1, e_2), n_5 = (e_2, r_1, e_1)$. Our predictor $f_{r_1}$ of positive links for relation $r_1$ can be trained in such
a way that it outputs the highest score for the true link and slightly lower scores for the negative links, i.e., $f_{r_i}(t_1) = 0.99, f_{r_i}(n_i) = 0.98$. Obviously,
$f_{r_i}(t_1)$ would be ranked above all $f_{r_i}(n_i)$, yielding a perfect mean rank $= 1$ score. However, if we now wanted to use $f_{r_i}$ as our binary predictor $P_{r_i}(e_i, r_k, e_j) = 1$ if $f_{r_i}(e_i, r_k, e_j) > 0.5$, and $P_{r_i}(e_i, r_k, e_j) = 0$ if $f_{r_i}(e_i, r_k, e_j) \le 0.5$, we would have a predictor $P_{r_i}$ which predicts all links to be \emph{true}. Obviously, we could have tuned (e.g., identify the suitable threshold from a few
samples) our predictor to have a threshold of $0.99$ (i.e., $\forall i, 0.5 < f_{r_i}(n_i) < 0.99$), but that would be a non-flexible predictor. In other words, the mean
rank measure may evaluate certain embeddings to be of good quality, while they may actually be bad for a specific link prediction task. We believe that the mean rank
measurement should be used to judge the quality of the embeddings for the \emph{reconstruction}~\cite{nickel_2017} the original knowledge graph that was used to train the embeddings, as opposed to predict new links. 




\subsubsection{Link prediction as binary classification task}

Mean rank based link prediction pipeline does not individuate the performance of entity embeddings to predict a specific relation, as it gives a scalar evaluation of
overall performance. Such a performance description is akin to describing the whole population distribution with its mean value only. In the bioinformatics field, link
prediction is a very important problem, and it is formulated as a binary classification task~\cite{alshahrani_2017}. This is different from the \emph{reconstruction}
setting. Instead of asking to which other entity among all possible entities $e_1, \ldots, e_n$ the entity $e_1$ is more probable to be connected with a link $r_1$, we
ask what is the probability of having a link $P((e_1, r_1, e_2) = 1) ?$ An example of $P((e_1, r_1, e_2) = 1)$ could be asking:

\begin{quote}
    What is the probability that the gene \emph{TRIM28} ($=e_1$) \textsc{has function} ($=r_i$) \emph{negative regulation of transcription by RNA polymerase II}
    ($=e_j$)? 
\end{quote}

Indeed, a predictor $P$ that ranks $(e_1, r_1, e_2)$ first among all other possible completions of $(e_1, r_1, x), \forall x \in KG, x \ne e_2$ is not a strong enough
link predictor. Apart from the fact that many of these potential negative examples would not even make sense biologically, i.e., $x \not \in \text{range(\textsc{has
function})}$, the predictor $P$ risks to accept too many false positives before it correctly classifies all the true positive examples. For the rest of this paper we focus on evaluation strategies of link prediction tasks for knowledge graphs, where link prediction is formulated as a binary classification problem.

\subsubsection{How much multi-relational are knowledge graphs?} \label{sec:multi-relationness}

Typically, in the statistical relation learning field the link prediction algorithms embed both entities and relations in the same embedding space, and then use
binary operators defined on these representations to represent a labelled link (i.e., a triple). For instance, given a function $\gamma$ that associates entities or
relations to its vector space embeddings, a predictor $f(e_1, r_1, e_2)$ might output the score by evaluating a standard Euclidean dot product like so $f(e_1, r_1, e_2) =
\langle \gamma(e_1) + \gamma(r_1), \gamma(e_2) \rangle$. And if $\gamma(r_1) \ne \gamma(r_2)$ then we expect $\langle \gamma(e_1) + \gamma(r_1), \gamma(e_2) \rangle$ and
$\langle \gamma(e_1) + \gamma(r_2), \gamma(e_2) \rangle$ to output very different scores, disambiguating thus links with relation $r_1$ or $r_2$ between the two entities
$e_1$ and $e_2$. 

\begin{figure}[!h]
    \begin{center}
        \includegraphics[width=.5\textwidth]{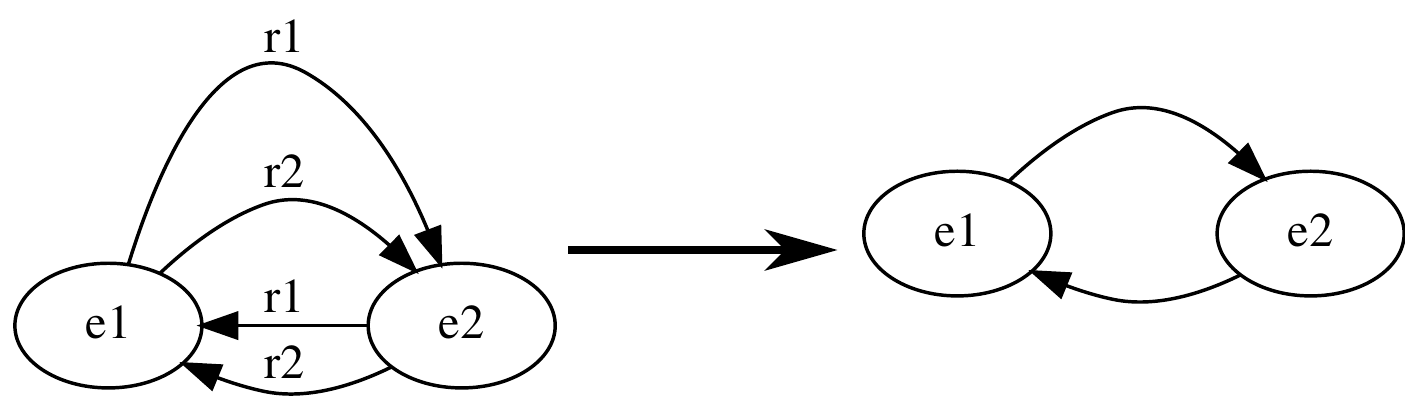}
    \end{center}
    \caption{\label{fig:flat-graph} Flattening a Knowledge Graph.}
\end{figure}

One might ask how it is possible to construct link predictors by only considering the embeddings of the entities? Can we \emph{flatten} the KG, by introducing the
restriction that each pair of entities must have at most one unlabelled link (Figure~\ref{fig:flat-graph}), and come up with a strategy of disambiguating links with
different relations? To better demonstrate the argument, below in Table~\ref{tab:wn11-stats} we provide the statistics of pairwise count of links in the WN11 Knowledge
Graph (original WordNet dataset~\cite{miller_1995} brought down to 11 relations as in~\cite{bordes_2013}), we can see that only $0.133 \%$ of all pairs are connected with
more than one link. In other words the \emph{multi-relationness} of this graph is really low. 

\begin{table}
\centering 
\scriptsize
    \begin{tabular}{llrr}
\toprule
                      relation  & \# links             & \# sources          & \# targets  \\
\midrule
 derivationally related form    & 31867               & 16737               & 16737       \\
                    hypernym    & 37221               & 36347               & 9795        \\
     member of domain region    & 983                 & 118                 & 925         \\
      synset domain topic of    & 3335                & 3170                & 313         \\
      member of domain usage    & 675                 & 25                  & 635         \\
              member meronym    & 7928                & 3238                & 7858        \\
                  similar to    & 86                  & 82                  & 82          \\
                    has part    & 5142                & 2062                & 4223        \\
                    also see    & 1396                & 727                 & 828         \\
                  verb group    & 1220                & 1038                & 1038        \\
           instance hypernym    & 3150                & 2622                & 419         \\
\midrule
                                & $a=$\#pairs (multi) & $b=$\#pairs (total) & $a/b$ (\%)  \\
                                & 124                 & 93003               & 0.133\%     \\
\bottomrule
\end{tabular}

    \caption{\label{tab:wn11-stats}Statistics on the connectivity of the entities per relation in WN11 knowledge graph.}
\end{table}

Perhaps not directly, but similarly, some works have considered embedding the entities without the relations~\cite{alshahrani_2017,nickel_2017} (although the
potential pitfalls of flattening the links are not raised there). In~\cite{nickel_2017} the link prediction of a hierarchical transitive closure is proposed to be treated
without learning an embedding for the hierarchical relation (e.g., \textsc{hypernym}). In the bioinformatics domain a similar idea of learning embeddings separately for
each relation $r_i$ is proposed, further enhanced by considering binary classifiers operating on the learned embeddings to predict links of type
$r_i$~\cite{alshahrani_2017}.


\subsubsection{Contribution of this work}

We investigate empirically the potential of training neural embeddings \emph{globally} for the entire graph, as opposed to training embeddings \emph{locally} for a
specific relation $r_i$. This is different to what has been previously done~\cite{nickel_2017,alshahrani_2017}, and opens new ways of assessing the quality of the neural
embeddings. By comparing, combining and extending different methodologies for link prediction on graph-based data coming from different domains, we present a unified
methodology for the quality evaluation of neural embeddings for link prediction tasks for knowledge graphs. The link prediction problem is treated here as a binary
classification task on entity embeddings. The evaluation steps required for an effective assessment of the quality of the knowledge graph embeddings are formalized. Our
evaluation pipeline is made open source~\cite{github_repo}, and with this we aim to draw more attention of the community towards an important issue of
transparency and reproducibility of the results. 


%
\section{Methods}
\subsubsection{Preparation of the datasets for neural embedding training} \label{sec:evaluation-embeddings}

The preparation of datasets for neural embedding training is inspired by the methodology presented in~\cite{alshahrani_2017}. In the following we present our generalized
approach to this problem, as we think it is crucial for the transparent and reproducible evaluation pipeline, and has not been detailed enough in other work.  In addition
to \emph{local} retained graphs,  where only triples of a specified relation $r_i$ were removed (as used in~\cite{alshahrani_2017}), we also consider \emph{global}
retained graphs for all relation $\forall r_i \in KG$.

\begin{figure}[!h]
    \begin{center}
        \includegraphics[width=.9\textwidth]{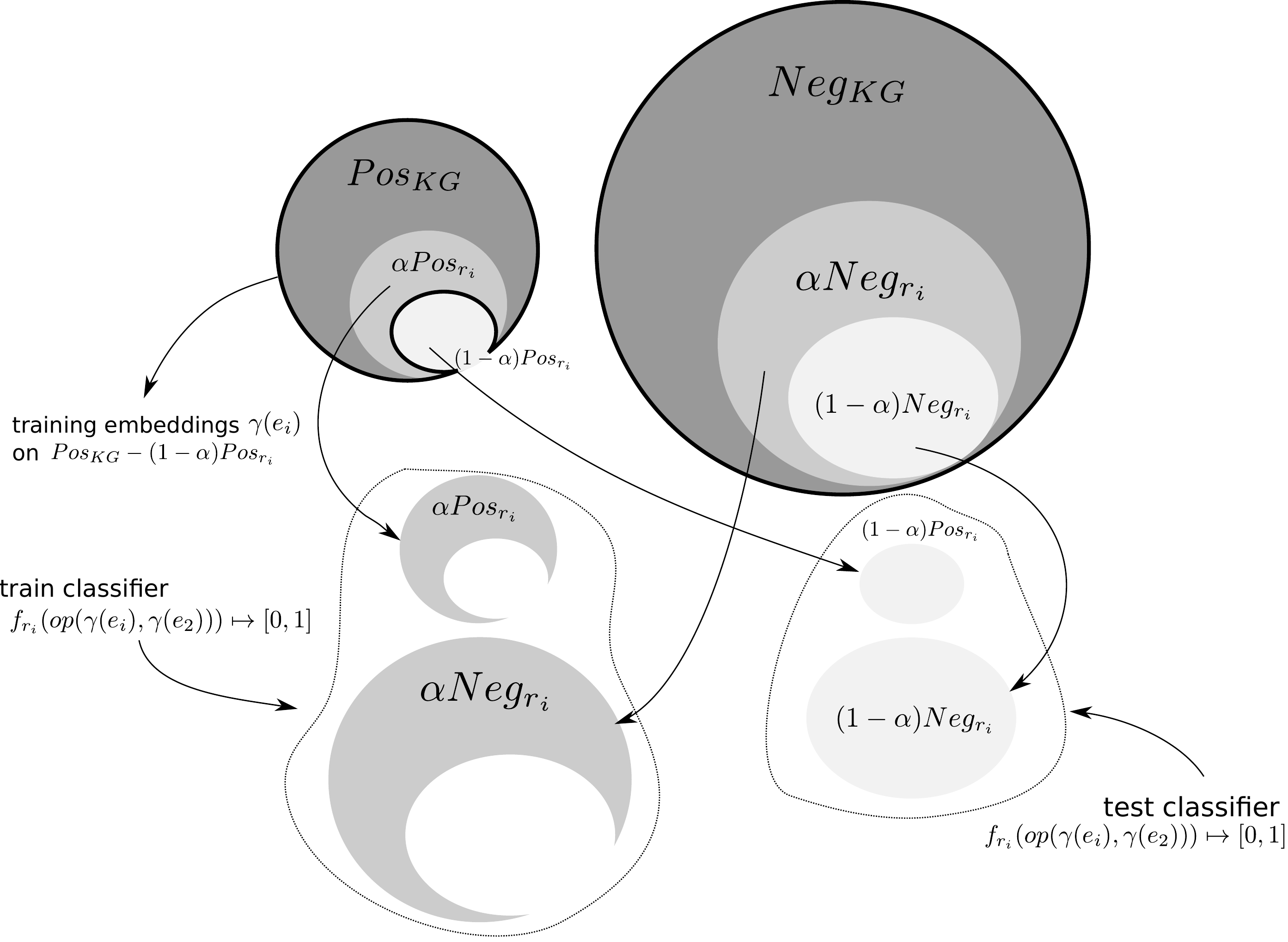}
    \end{center}
    \caption{\label{fig:evaluation} Schematic representation of the pipeline for the evaluation of the embeddings. $Neg_{KG}$ and its derivations (e.g., $\alpha
    Neg_{r_i}$) appear bigger visually to indicate that the elements are sampled from a much bigger set of all possible negative links.}
\end{figure}

Let $Pos_{KG}$ denote all the existing links in the KG, i.e., triples $(e_i, r_i, e_j) \in KG$, and let $Neg_{KG}$ denote the non-existing links $(\bar{e_i}, r_i,
\bar{e_j}) \not \in KG$, with the restrictions that $|Pos_{KG}| = |Neg_{KG}|$ (number of elements is equal), and $\bar{e_i} \in \textsc{domain}(r_i)$, and $\bar{e_j} \in
\textsc{range}(r_i)$, as in~\cite{alshahrani_2017}. The former restriction ($|Pos_{KG}| = |Neg_{KG}|$) ensures that we do not have imbalanced classes for the binary
prediction (see~\cite{garciagasulla_2016} for more details), we therefore sample negatives at random with the sample size equal to the number of positive examples.
Potentially, the set of all possible negative examples is much bigger than the set of all positive examples (because we are enumerating many more possible connections in
the graph, excluding the existing ones). The latter restriction on the domain and the range of a relation fixes attention to the most probable and semantically consistent
negative links. Then, $Pos_{KG}$ forms the set of all \emph{positive} examples, analogously, $Neg_{KG}$ -- set of sampled \emph{negative} examples (see
Section~\ref{sec:intro-lp}). To test the quality of the embeddings and their ability to classify examples for a specific relation, we split for each considered relation
$r_i$ the set of positive and negative links of type $r_i$ ($Pos_{r_i}, Neg_{r_i}$) in a given train/test ratio $\alpha \in [0, 1]$ ($\alpha = 0.8$
in~\cite{alshahrani_2017}). To simplify the notation, we let $\alpha Pos_{r_i}$ represent the train split of all positive examples for links with type $r_i$, and $(1 -
\alpha) Pos_{r_i}$ represent the test split of all positive examples.  Similarly, for the negative examples (i.e., train split $\alpha Neg_{r_i}$ and test split $(1 -
\alpha) Neg_{r_i}$). Essentially, when we say $(\alpha = 0.8) Pos_{r_i}$ we mean the set consisting of $80 \%$ of positive examples of type $r_i$ (we hope that the abuse
of notation will increase comprehension).  Figure~\ref{fig:evaluation} employs a set-theoretic depiction of the sets of all positive and negative links, as well as their
subsets for links restricted to a specific type $r_i$, and divisions into train and test splits.

By treating the problem of evaluation of the quality of the embeddings in this set-theoretic approach, we now define the following \emph{datasets}:

\begin{enumerate}
    \item $Pos_{KG} - (1-\alpha) Pos_{r_i}$ a \emph{local} retained graph on $r_i$ -- training corpus for unsupervised learning of local to $r_i$ entity embeddings $\gamma_{r_i}(e_i)$ (in
        Figure~\ref{fig:evaluation} this set is demarcated with bold contour in the upper left corner),
    \item $Pos_{KG} - \bigcup_{r_i} (1-\alpha) Pos_{r_i}$ a \emph{global} retained graph on all relations $r_i$ -- training corpus for unsupervised learning of global entity embeddings $\gamma(e_i)$,
    \item $\forall r_i, \alpha Pos_{r_i} \bigcup \alpha Neg_{r_i}$ -- train examples for the binary classifier on $r_i$,
    \item $\forall r_i,(1-\alpha) Pos_{r_i} \bigcup (1 - \alpha) Neg_{r_i}$ -- test examples for the binary classifier $r_i$.
\end{enumerate}

We also note the following properties, which must hold and serve as validation criteria for the generation of train and test data. In particular, 

\begin{enumerate}
    \item $Pos_{KG} - (1-\alpha) Pos_{r_i} \subseteq Pos_{KG}, Pos_{KG} - \bigcup_{r_i} (1-\alpha) Pos_{r_i} \subseteq Pos_{KG}$ local and global retained graphs must be
        subgraphs of full graph,
    \item $Pos_{KG} - \bigcup_{r_i} \alpha Pos_{r_i} = \bigcup_{r_i} (1-\alpha) Pos_{r_i}$ the difference between the
        full and retained global graphs are the positive links, which we use in the test set,
        i.e., the embeddings will be used to predict these positive links,
    \item $\forall r_i, \alpha Pos_{r_i} \bigcap (1 - \alpha) Pos_{r_i} = \varnothing, \alpha Neg_{r_i} \bigcap (1 - \alpha) Neg_{r_i} = \varnothing$ there should be no
        link shared between the train and test sets,
    \item $\forall r_i \forall x_i, \bar{x_i} \in Neg_{r_i}, \bar{x_i} \not \in KG$ all generated negative links do not exist in the original full graph.
\end{enumerate}

The last two properties reflect what is usually done in the literature during the generation of negative links, and for this work we also conform to these two properties.
However, we elaborate more on the issue, where the negative examples set generation is disjoint from the positive examples set in Section~\ref{sec:discussion}.

\subsubsection{Training neural embeddings} In this work we employ a fast and scalable unsupervised neural embedding model~\cite{agibetov_2018}, which aims at learning
\emph{entity embeddings}, each of which is described by a set of discrete \emph{features} (bag-of-features) coming from a fixed-length dictionary. The model is trained by
assigning a $d$-dimensional vector to each of the discrete features in the set that we want to embed directly. Ultimately, the look-up matrix (the matrix of embeddings -
latent vectors) is learned by minimizing the following loss function $$ \sum_{(a, b) \in E^+, b^- \in E^-} L^{batch} (sim(a, b), sim(a, b_1^-), \ldots, sim(a, b_k^-)).
$$ In this loss function, we need to indicate the generator of positive entry pairs $(a, b) \in E^+$, and the generator of negative entities $b_i^- \in E^-$, similar to
the $k$-negative sampling strategy proposed by Mikolov et al.~\cite{mikolov_2013}. In our setting, for each entity $e_i$ in the knowledge graph, $E^+$ is the generator of
entities $(a=e_i, b=e_j)$ from the local $Pos_{KG} - (1-\alpha) Pos_{r_i}$ or global $Pos_{KG} - \bigcup_{r_i} (1-\alpha) Pos_{r_i}$ retained graphs, and $E^-$ is the
generator of a negative entity $b^-=\bar{e_j}$, such that $(e_i, \bar{e_j}) \not \in KG$. $L^{batch}$ denotes that we minimize the objective function for the small
subsets of elements drawn from $E^+$ and $E^-$ (mini-batching). The similarity function $sim$ is task-dependent and should operate on $d$-dimensional vector
representations of the entities, in our case we use the standard Euclidean dot product. This model (and implicitly the embeddings are) is trained with the StarSpace
toolkit~\cite{wu_2017}. The aforementioned embedding scheme is different from a multi-relational knowledge graph embedding task, since we do not require the explicit
embeddings for the relations (Section~\ref{sec:multi-relationness}). 

Please note that since we are learning embeddings for the entities from the retained graphs (some links are excluded), the algorithm may \emph{miss} to learn an
embedding for an entity. That is, suppose that during the generation of the retained graph all connectivities $\forall x, (e_k, x)\in KG$ of an entity $e_k$ are not
assigned to the retained graph $\forall x, (e_k, x) \not \in Pos_{KG} - (1-\alpha) Pos_{r_i}$, then the algorithm will not learn an embedding $\gamma(e_k)$, which will
lead us to a situation where all the pairs $\forall x, (e_k, x) \in KG$ will be missing during training or testing of the binary classifier (depending whether these pairs
are assigned to to the train or test sets). Obviously, the amount of possible missing embeddings is inversely proportionate to the $\alpha$ parameter, i.e., the more
information we include during the embedding learning phase, the fewer embeddings will be missed.


\subsubsection{Binary operators for link representation}

Based on the embeddings of the nodes of the graph, we can come up with different ways of representing a link between an entity $e_i$ and $e_j$.  This is usually achieved
with a binary operator $op$ that combines entitiy embeddings representations $\gamma(e_i), \gamma(e_j)$ into one single representation of the link $(e_i, r_i, e_j)$.
Popular choices for this operator include operations that preserve the original $d$ dimension of the entity embeddings to represent links (e.g., element-wise sum or
mean~\cite{grover_2016}), as well the operations that combine entitiy embeddings, such as concatenation~\cite{agibetov_2018}. The definitions of these operators are given
in Table~\ref{tab:def-operators}; we use them in our experiments and evaluation.

\begin{table}
\centering 
\begin{adjustbox}{width=1\textwidth}
    \scriptsize
    \setlength\tabcolsep{10pt}
    \begin{tabular}{llp{.3\textwidth}}
\toprule
Operator $op$ definition & Link representation & Comments \\
\midrule
$sum(\gamma(e_i), \gamma(e_j)): \mathbb{R}^n \times \mathbb{R}^n \mapsto \mathbb{R}^n$ & $sum(\gamma(e_i), \gamma(e_j)) = \begin{bmatrix} \gamma(e_i)_i + \gamma(e_j)_i \\ \vdots \\ \gamma(e_i)_n + \gamma(e_j)_n \end{bmatrix}$ & element-wise sum of the two vectors \\
$mean(\gamma(e_i), \gamma(e_j)): \mathbb{R}^n \times \mathbb{R}^n \mapsto \mathbb{R}^n$ & $mean(\gamma(e_i), \gamma(e_j)) = \begin{bmatrix} (\gamma(e_i)_i + \gamma(e_j))/2 \\ \vdots \\ \gamma(e_i)_n + \gamma(e_j)_n)/2 \end{bmatrix}$ & element-wise arithmetic mean of the two vectors \\
$concat(\gamma(e_i), \gamma(e_j)): \mathbb{R}^n \times \mathbb{R}^n \mapsto \mathbb{R}^{2n}$ & $concat(\gamma(e_i), \gamma(e_j)) = \begin{bmatrix} \gamma(e_i) \\ \gamma(e_j) \end{bmatrix}$ & concatenation of the two vectors (double the size of original embedding) \\
\bottomrule
\end{tabular}

\end{adjustbox}
    \caption{\label{tab:def-operators}Binary operators for link representation from the entity embeddings $\gamma(e_i), \gamma(e_j)$.}
\end{table}




\subsubsection{Repeated random sub-sampling validation}

To quantify confidence in the trained embeddings, we perform the repeated random sub-sampling validation for each classifier $f_{r_i}$. That is, for each relation
$r_i$ we generate $k$ times: retained graph $Pos_{KG} - (1-\alpha) Pos_{r_i}$ corpus for unsupervised learning of entity embeddings $\gamma_{r_i}(e_i)$) and train $\alpha
Pos_{r_i} \bigcup \alpha Neg_{r_i}$ and test $(1-\alpha) Pos_{r_i} \bigcup (1 - \alpha) Neg_{r_i}$ splits of positive and negative examples. Link prediction is then
treated as a binary classification task with a logistic regression classifier $f_{r_i}(op(\gamma(e_i), \gamma(e_j))) \mapsto [0, 1]$ defined on the link representation
produced by the binary operator (e.g., $op=sum$). The performance of the classifier is measured with the standard performance measurement based on false positive rate to
true positive rate curves. We, in particular, report the F1 score (i.e., F-measure with equal weights for precision and recall).

\section{Results}
%
In this section we report on the possibility of training the embeddings $\forall e_i \in KG, \gamma(e_i)$ once on the retained graph $Pos_{KG} - \bigcup_{r_i} Pos_{r_i}$
on all relations $r_i$, and evaluating it separately on all train $\alpha Pos_{r_i}$ and test examples $(1-\alpha) Pos_{r_i}$ for each relation $r_i$. 
We evaluate the quality of local and global embeddings and study the influence of the choice of the binary operator (e.g., sum, concatenation) used in the logistic
regression classifier $f_{r_i}(op(\gamma (e_i), \gamma (e_j)))$, as well as the size (controlled by $\alpha$) of the retained local and global graphs, applied to the
WN11 knowledge graph. All of our results are presented as averages of 10 repeated random sub-sampling validations
(Section~\ref{sec:evaluation-embeddings}). We therefore report mean F-measure scores and their standard deviations. All embeddings are trained with fixed hyperparameters:
embedding size is set to $d=50$ and number of epochs is 10. The neural embeddings are trained with the StarSpace toolkit~\cite{wu_2017}. Classification results are
obtained with the scikit Python library~\cite{pedregosa_2011}, grouping of data and their statistical analysis are performed with Pandas~\cite{mckinney-proc-scipy-2010}.
All of our experiments were performed on a modern desktop PC with a quad core Intel i7 CPU (clocked at 4GHz) and 32 Gb of RAM.

\begin{table}
\centering 
\begin{adjustbox}{width=1\textwidth}
\setlength\tabcolsep{0pt}
    \begin{tabular}{lSSSSSS}
\toprule
    $\alpha = 0.2$ & & & & & &\\ 
    \multirow{3}{*}[0.5em]{relation}   &
    \multicolumn{3}{c}{{Local}}          &
    \multicolumn{3}{c}{{Global}}                                                                                                                             \\
                                           & {sum}            & {concatenation}  & {mean}          & {sum}           & {concatenation}  & {mean}          \\
\midrule
 member of domain region                    & 0.94  \pm 0.02  & 0.94 \pm 0.02  & 0.91 \pm 0.05  & 0.91 \pm 0.06  & 0.94           & 0.73 \pm 0.14   \\
 member meronym                             & 0.96  \pm 0.02  & 0.98 \pm 0.01  & 0.95 \pm 0.02  & 0.98 \pm 0.02  & 0.99           & 0.96 \pm 0.04  \\
    similar to                                 & 0.93  \pm 0.02  & 0.95\textsuperscript{*}           & 0.91 \pm 0.07  & 0.66 \pm 0.17  & 0.7  \pm 0.21  & 0.55 \pm 0.03  \\
 instance hypernym                          & 0.66  \pm 0.03  & 0.67 \pm 0.04  & 0.66 \pm 0.03  & 0.61 \pm 0.07  & 0.63 \pm 0.06  & 0.6  \pm 0.06  \\
 has part                                   & 0.9   \pm 0.03  & 0.92 \pm 0.04  & 0.9  \pm 0.03  & 0.88 \pm 0.06  & 0.91 \pm 0.08  & 0.8  \pm 0.09  \\
 derivationally related form                & 0.76  \pm 0.01  & 0.77           & 0.76 \pm 0.01  & 0.83 \pm 0.15  & 0.84 \pm 0.15  & 0.83 \pm 0.15   \\
 also see                                   & 0.82  \pm 0.09  & 0.83 \pm 0.09  & 0.82 \pm 0.09  & 0.83 \pm 0.13  & 0.84 \pm 0.14  & 0.81 \pm 0.11   \\
 verb group                                 & 0.89  \pm 0.02  & 0.91           & 0.88 \pm 0.02  & 0.88 \pm 0.04  & 0.9  \pm 0.04  & 0.85 \pm 0.07  \\
 member of domain usage                     & 0.91  \pm 0.03  & 0.92 \pm 0.03  & 0.88 \pm 0.04  & 0.88 \pm 0.08  & 0.9  \pm 0.08  & 0.61 \pm 0.05  \\
 hypernym                                   & 0.72  \pm 0.06  & 0.72 \pm 0.06  & 0.71 \pm 0.06  & 0.71 \pm 0.14  & 0.72 \pm 0.15  & 0.71 \pm 0.14   \\
 synset domain topic of                     & 0.65  \pm 0.01  & 0.68 \pm 0.01  & 0.64 \pm 0.01  & 0.64 \pm 0.06  & 0.65 \pm 0.05  & 0.64 \pm 0.06  \\
\midrule
    averaged values                         & 0.82 \pm 0.11   & 0.84 \pm 0.03  & 0.81 \pm 0.1   & 0.8 \pm 0.12   & 0.82 \pm 0.12  & 0.73 \pm 0.12   \\
\midrule
    $\alpha = 0.5$          &                 &                &                &                &                & \\
                                            &                 &                &                &                &                & \\
\midrule
 member of domain region                    & 0.93 \pm 0.02   & 0.93 \pm 0.02  & 0.9  \pm 0.03  & 0.92 \pm 0.01  & 0.92 \pm 0.02  & 0.9  \pm 0.02   \\
 member meronym                             & 0.97 \pm 0.01   & 0.98 \pm 0.01  & 0.97 \pm 0.01  & 0.97 \pm 0.02  & 0.98 \pm 0.01  & 0.96 \pm 0.02  \\
 similar to                                 & 0.9  \pm 0.03   & 0.92 \pm 0.01  & 0.9  \pm 0.03  & 0.89 \pm 0.04  & 0.92 \pm 0.01  & 0.9  \pm 0.03  \\
 instance hypernym                          & 0.65 \pm 0.04   & 0.66 \pm 0.03  & 0.65 \pm 0.05  & 0.62 \pm 0.04  & 0.63 \pm 0.06  & 0.61 \pm 0.04  \\
 has part                                   & 0.85 \pm 0.02   & 0.85           & 0.85 \pm 0.02  & 0.83 \pm 0.05  & 0.83 \pm 0.06  & 0.83 \pm 0.05  \\
 derivationally related form                & 0.67 \pm 0.01   & 0.68           & 0.67 \pm 0.01  & 0.68 \pm 0.12  & 0.7  \pm 0.11  & 0.68 \pm 0.12   \\
 also see                                   & 0.76 \pm 0.13   & 0.77 \pm 0.13  & 0.76 \pm 0.13  & 0.76 \pm 0.13  & 0.76 \pm 0.13  & 0.76 \pm 0.13   \\
 verb group                                 & 0.85 \pm 0.01   & 0.87           & 0.85 \pm 0.01  & 0.85           & 0.86           & 0.85   \\
 member of domain usage                     & 0.86 \pm 0.08   & 0.86 \pm 0.08  & 0.85 \pm 0.08  & 0.85 \pm 0.1   & 0.85 \pm 0.1   & 0.85 \pm 0.09  \\
 hypernym                                   & 0.64 \pm 0.04   & 0.65 \pm 0.04  & 0.64 \pm 0.04  & 0.61 \pm 0.12  & 0.63 \pm 0.13  & 0.61 \pm 0.12   \\
 synset domain topic of                     & 0.64 \pm 0.02   & 0.68 \pm 0.03  & 0.64 \pm 0.02  & 0.62 \pm 0.04  & 0.67 \pm 0.06  & 0.62 \pm 0.05   \\
\midrule
    averaged values                         & 0.79 \pm 0.12   & 0.8 \pm 0.11   & 0.78 \pm 0.12  & 0.78 \pm 0.13  & 0.79 \pm 0.12  & 0.77 \pm 0.13   \\
\midrule
    $\alpha = 0.8$          &                 &                &                &                &                & \\
                                            &                 &                &                &                &                & \\
\midrule
 member of domain region                    & 0.85 \pm 0.01   & 0.8  \pm 0.09  & 0.84 \pm 0.01  & 0.83 \pm 0.01  & 0.8  \pm 0.11  & 0.82 \pm 0.02  \\
 member meronym                             & 0.96 \pm 0.01   & 0.97 \pm 0.01  & 0.96 \pm 0.01  & 0.96 \pm 0.01  & 0.97 \pm 0.01  & 0.96 \pm 0.01  \\
 similar to                                 & 0.83 \pm 0.02   & 0.84 \pm 0.01  & 0.83 \pm 0.03  & 0.82 \pm 0.05  & 0.84 \pm 0.02  & 0.81 \pm 0.05  \\
 instance hypernym                          & 0.65 \pm 0.05   & 0.69 \pm 0.09  & 0.65 \pm 0.05  & 0.66 \pm 0.06  & 0.67 \pm 0.07  & 0.66 \pm 0.06  \\
 has part                                   & 0.69 \pm 0.01   & 0.69           & 0.7  \pm 0.04  & 0.68 \pm 0.02  & 0.67           & 0.69 \pm 0.05  \\
 derivationally related form                & 0.92 \pm 0.04   & 0.99           & 0.92 \pm 0.04  & 0.91 \pm 0.05  & 0.99           & 0.91 \pm 0.05  \\
 also see                                   & 0.94 \pm 0.04   & 0.99           & 0.92 \pm 0.05  & 0.92 \pm 0.06  & 0.99           & 0.89 \pm 0.08  \\
 verb group                                 & 0.71 \pm 0.01   & 0.72           & 0.71 \pm 0.01  & 0.71 \pm 0.01  & 0.72           & 0.7  \pm 0.01  \\
 member of domain usage                     & 0.8  \pm 0.08   & 0.85 \pm 0.01  & 0.79 \pm 0.08  & 0.79 \pm 0.1   & 0.84 \pm 0.01  & 0.78 \pm 0.1    \\
 hypernym                                   & 0.6  \pm 0.02   & 0.77 \pm 0.12  & 0.6  \pm 0.02  & 0.57 \pm 0.01  & 0.69 \pm 0.09  & 0.57 \pm 0.01  \\
 synset domain topic of                     & 0.65 \pm 0.03   & 0.7  \pm 0.04  & 0.63 \pm 0.03  & 0.63 \pm 0.04  & 0.72 \pm 0.08  & 0.62 \pm 0.04  \\
\midrule
    averaged values                         & 0.78 \pm 0.12   & 0.82 \pm 0.12  & 0.77 \pm 0.12  & 0.77 \pm 0.13  & 0.8 \pm 0.12   & 0.76 \pm 0.12   \\
\bottomrule
\end{tabular}

\end{adjustbox}
    \caption{\label{tab:wn11-operators}Comparison of the F-measure scores of the binary classifiers that use different binary operators for link representation. The
    scores are given for local and global evaluation of neural embeddings, with varying amount of available information about the connectivity of the WN11 knowledge
    graph. Scores presented without confidence intervals indicate small variance (i.e., $\sigma < 0.01$).}
\end{table}
 
Table~\ref{tab:wn11-operators} regroups averaged cross-validation scores of the $f_{r_i}$ binary classifiers per relations $r_i$. These scores are split per increasing amount of information (controlled by $\alpha$) which is available during the unsupervised learning phase of the
neural embeddings. This models the realistic scenario where links represent an ever growing knowledge about the domain, and where we want to predict new links that might
emerge in future. Overall, the unsupervised training phase seem to be quiet robust to limited amounts of information (i.e., training embeddings with only 20\% of links
vs. training with 80\% of available links), as average F-measure score for all relations $r_i$ seem to be affected only slightly in both local and global settings.  We
can notice that the concatenation binary operator outperforms other link representations, and all its predictions lie within the $0.79-0.84$ range. This observation might
be caused by the fact that the output of the concatenation operator has twice the amount of dimensions to represent information (i.e., $concat(\gamma(e_i), \gamma(e_j))
\in \mathbb{R}^{2d} \text{ vs. } sum(\gamma(e_i), \gamma(e_j)) \in \mathbb{R}^{d}$). Besides, unlike sum and mean, it naturally encodes \emph{directionality} of links
(i.e., assymetric relations $concat(\gamma(e_i), \gamma(e_j)) \ne concat(\gamma(e_j), \gamma(e_i))$). 

Depending on the connectivity of the knowledge graph, retaining some ratio (i.e., $1 -\alpha$) of available links may have severe consequences on the number of missing
embeddings for the entities with very low incoming and outgoing links.  In Table~\ref{tab:wn11-global-local} we group F-measure scores for the concatenation operator, and
additionally, report the percentage of missing embeddings in both train and test dataset splits for each relation $r_i$. As expected, pre-training the embeddings on only
20\% of all relations $(Pos_{KG} - \bigcup_{r_i} Pos_{r_i})$ will miss many entities compared to the local setting $(Pos_{KG} - Pos_{r_i})$, where we train on 20\% only
for a specific relation $r_i$ and on 100\% of links for $r_j \ne r_i$.  29.38 \% vs. 3.7 \% of missed training examples for global and local settings respectively,
analogously, 59.47 \% vs. 7.8 \% for test examples. This makes a big difference in confidence of our link prediction binary classifier $f_{r_i}$ trained locally or
globally, even if the final F-measure scores are quiet comparable. However, if we consider $\alpha=0.8$ then the percentage of missing examples for train ($0.56 \%$) and
test ($5.69 \%$) splits for global setting are tolerable, with $0.13 \%$ and $2.19 \%$ for the local approach respectively.  Obvious advantage of the global approach is
scalability in both time and space. We train and store only one neural model (embedding vectors are stored implicitly in the weight matrix of the hidden layer) for the
global approach, and we need to train and store embeddings locally for as many models as there are relations in the knowledge graph. For the WN11 KG, if we consider
$\alpha=0.8$, we need on average $\approx 25$ sec to train one global model, and we require on average $\approx 32$ seconds to train a model per relation for the local
approach (averaged over 10 repeated random sub-sampling validations). Since we have 11 relations, we thus need $\approx 25$ seconds vs. $\approx 352$ seconds for $d=50$ and 10 epochs. The time needed to
train these models will obviously grow as we increase the embedding dimension and the number of epochs. Spacewise, the global approach needs $21$ Mb and the local $242$
Mb, as in the case of time complexity, the memory needed to store bigger models ($d > 50$) will increase.

%

\begin{table}
\centering 
\begin{adjustbox}{width=1\textwidth}
\setlength\tabcolsep{0pt}
\begin{tabular}{lSSSSSS}
\toprule
$\alpha = 0.2$ & & & & & & \\ 
    \multirow{3}{*}[0.5em]{relation}   &
    \multicolumn{3}{c}{Local}          &
    \multicolumn{3}{c}{Global}                                                                                                                           \\
                                        & {F-measure}     & {train miss (\%)}  & {test miss(\%)}  & {F-measure}     & {train miss (\%)}  & {test miss (\%)}  \\
\midrule
 member of domain region                 & 0.94 \pm 0.02  & 2.8   \pm 0.78  & 5.5  \pm 0.39   & 0.94           & 30 \pm 1.3      & 64 \pm 1.4        \\
 member meronym                          & 0.98 \pm 0.01  & 0.08 \pm 0.05   & 0.17 \pm 0.02   & 0.99           & 38 \pm 13       & 75 \pm 13         \\
 similar to                              & 0.95           & 0.88  \pm 2     & 4    \pm 1.9    & 0.7  \pm 0.21  & 14 \pm 3.8      & 42 \pm 6.5        \\
 instance hypernym                       & 0.67 \pm 0.04  & 12    \pm 14    & 21   \pm 6.9    & 0.63 \pm 0.06  & 34 \pm 12       & 67 \pm 5.7        \\
 has part                                & 0.92 \pm 0.04  & 0.44  \pm 0.14  & 2.2  \pm 3.5    & 0.91 \pm 0.08  & 38 \pm 13       & 71 \pm 11         \\
 derivationally related form             & 0.77           & 0.89  \pm 0.08  & 2.8  \pm 0.14   & 0.84 \pm 0.15  & 26 \pm 16       & 50 \pm 9          \\
 also see                                & 0.83 \pm 0.09  & 1.9   \pm 0.43  & 12   \pm 10     & 0.84 \pm 0.14  & 32 \pm 19       & 53 \pm 12         \\
 verb group                              & 0.91           & 0.23  \pm 0.23  & 0.64 \pm 0.24   & 0.9  \pm 0.04  & 19 \pm 16       & 41 \pm 15         \\
 member of domain usage                  & 0.92 \pm 0.03  & 6.1   \pm 15    & 7.3  \pm 14     & 0.9  \pm 0.08  & 33 \pm 9        & 67 \pm 9.5        \\
 hypernym                                & 0.72 \pm 0.06  & 15    \pm 18    & 29   \pm 8.8    & 0.72 \pm 0.15  & 29 \pm 18       & 64 \pm 8.3        \\
 synset domain topic of                  & 0.68 \pm 0.01  & 0.16  \pm 0.12  & 1    \pm 0.19   & 0.65 \pm 0.05  & 31 \pm 16       & 59 \pm 7.7        \\
\midrule
averaged values                          & 0.84 \pm 0.03  & 3.7 \pm 5.2     & 7.8 \pm 9.4     & 0.82 \pm 0.12  & 29.38 \pm 7.21  & 59.47 \pm 11.3    \\
\midrule
%
%
$\alpha = 0.5$                           &                &                 &                 &                &                 & \\
                                         &                &                 &                 &                &                 & \\
%
 member of domain region                 & 0.93 \pm 0.02  & 1.6   \pm 0.31  & 4.6  \pm 0.43   & 0.92 \pm 0.02  & 9.9 \pm 0.64    & 28  \pm 1.7       \\
 member meronym                          & 0.98 \pm 0.01  & 0.04 \pm 0.02   & 0.13 \pm 0.05   & 0.98 \pm 0.01  & 12  \pm 13      & 31  \pm 13        \\
 similar to                              & 0.92 \pm 0.01  & 0.58  \pm 0.82  & 2.2  \pm 2.4    & 0.92 \pm 0.01  & 1.1 \pm 0.66    & 7.8 \pm 4.6       \\
 instance hypernym                       & 0.66 \pm 0.03  & 4.8   \pm 8.2   & 15   \pm 5.8    & 0.63 \pm 0.06  & 11  \pm 8.8     & 34  \pm 6.9       \\
 has part                                & 0.85           & 0.22  \pm 0.05  & 0.8  \pm 0.17   & 0.83 \pm 0.06  & 12  \pm 13      & 28  \pm 9.4       \\
 derivationally related form             & 0.68           & 0.25  \pm 0.02  & 0.89 \pm 0.1    & 0.7  \pm 0.11  & 7.4 \pm 15      & 13  \pm 14        \\
 also see                                & 0.77 \pm 0.13  & 5.4   \pm 16    & 11   \pm 17     & 0.76 \pm 0.13  & 7   \pm 15      & 16  \pm 17        \\
 verb group                              & 0.87           & 0.06 \pm 0.07   & 0.21 \pm 0.17   & 0.86           & 1.3 \pm 0.25    & 5.8 \pm 1         \\
 member of domain usage                  & 0.86 \pm 0.08  & 0.65  \pm 0.26  & 2    \pm 0.6    & 0.85 \pm 0.1   & 9.6 \pm 1       & 30  \pm 5.6       \\
 hypernym                                & 0.65 \pm 0.04  & 4.4   \pm 7.2   & 14   \pm 6.2    & 0.63 \pm 0.13  & 5.4 \pm 7       & 26  \pm 14        \\
 synset domain topic of                  & 0.68 \pm 0.03  & 1.9   \pm 4     & 0.76 \pm 0.07   & 0.67 \pm 0.06  & 8.1 \pm 4.3     & 21  \pm 4.7       \\
\midrule
    averaged values                      & 0.8 \pm 0.11   & 1.8 \pm 2       & 4.67 \pm 5.73   & 0.79 \pm 0.12  & 7.75 \pm 3.87   & 21.84 \pm 9.73    \\
\midrule
%
%
$\alpha = 0.8$                           &                &                 &                 &                &                 & \\
                                         &                &                 &                 &                &                 & \\
 member of domain usage                  & 0.8  \pm 0.09  & 0.26  \pm 0.14  & 1.9   \pm 0.73  & 0.8  \pm 0.11  & 1.4   \pm 0.25  & 8.3  \pm 1.9      \\
 member meronym                          & 0.97 \pm 0.01  & 0.01            & 0.06 \pm 0.04   & 0.97 \pm 0.01  & 1.1   \pm 0.08  & 7    \pm 0.59     \\
 similar to                              & 0.84 \pm 0.01  & 0               & 0               & 0.84 \pm 0.02  & 0               & 0                 \\
 instance hypernym                       & 0.69 \pm 0.09  & 0.01 \pm 0.04   & 8.7   \pm 5.5   & 0.67 \pm 0.07  & 0.3   \pm 0.17  & 12   \pm 5.5      \\
 has part                                & 0.69           & 0.08 \pm 0.03   & 0.53  \pm 0.19  & 0.67           & 1     \pm 0.11  & 6.1  \pm 0.59     \\
 derivationally related form             & 0.99           & 0.03 \pm 0.01   & 0.25  \pm 0.07  & 0.99           & 0.16  \pm 0.03  & 1    \pm 0.15     \\
 also see                                & 0.99           & 0.09 \pm 0.08   & 0.68  \pm 0.5   & 0.99           & 0.22  \pm 0.12  & 1.3  \pm 0.61     \\
 verb group                              & 0.72           & 0.01 \pm 0.02   & 0.12  \pm 0.2   & 0.72           & 0.09 \pm 0.08   & 0.76 \pm 0.61     \\
 member of domain region                 & 0.85 \pm 0.01  & 0.64  \pm 0.14  & 3.2   \pm 0.55  & 0.84 \pm 0.01  & 1.7   \pm 0.28  & 9.2  \pm 1.2      \\
 hypernym                                & 0.77 \pm 0.12  & 0.31  \pm 0.02  & 8.2   \pm 0.14  & 0.69 \pm 0.09  & 0.34  \pm 0.02  & 10   \pm 0.17     \\
 synset domain topic of                  & 0.7  \pm 0.04  & 0               & 0.57  \pm 0.18  & 0.72 \pm 0.08  & 0.17  \pm 0.2   & 6    \pm 5.6      \\
\midrule
averaged values                          & 0.82 \pm 0.12  & 0.13 \pm 0.19   & 2.19 \pm 3.21   & 0.8 \pm 0.12   & 0.58 \pm 0.59   & 5.69 \pm 4.31     \\
\bottomrule
\end{tabular}

\end{adjustbox}
    \caption{\label{tab:wn11-global-local}F-measure scores for the binary classifiers that use concatenation operator for link representation, together with the
    percentage of missed embeddings due to limited available information on the connectivity of the WN11 knowledge graph. Scores presented
    without confidence intervals indicate small variance (i.e., $\sigma < 0.01$).}
\end{table}

\section{Discussion} \label{sec:discussion}
\subsubsection{Related work}
The most focused study of the link prediction problem for the large scale (unlabelled) graph-based data mining has been conducted in~\cite{garciagasulla_2016}, to the
best of our knowledge. The focus of that work is on negative sample generation, and the author emphasize that the link prediction problem is a hugely imbalanced binary
prediction task, where the number of negative samples is orders of magnitude higher than the number of positive examples. 
In the bioinformatics community Alshahrani et al.~\cite{alshahrani_2017} proposed to circumvent the problem of imbalanced classes for the binary classification problem by
considering negative links that have a biological meaning, truncating thus many potential negative links that are highly improbable biologically. They do this by
restricting all the negative links to have the same domain and range as the positive links (i.e., they do not consider highly improbable links of type $gene_i$
\textsc{has function} $drug_j, drug_j \not \in range(\text{\textsc{has function}})$). Link prediction for the entire knowledge graph is then treated as a set of binary
classification tasks (one for each relation). Both of these works agree that link prediction should be treated as a binary classification task. 
Some works have focused their attention on the strategies for data splitting, producing biased train and test examples, such that the implicit information from the test
set may leak into the train set~\cite{toutanova_2016,dettmers_2017}. In~\cite{dettmers_2017}, authors show that the random splits for the common knowledge graph
evaluation benchmarks (Wordnet~\cite{miller_1995} and Freebase~\cite{bollacker_2008}) may bias the classification results for the symmetric relations. Solutions to
unbiased evaluations include curated data splits where no such information leakage is present. Kadlec et al.~\cite{kadlec_2017} have mentioned that fair optimization of
hyperparameters for competing approaches should be considered, as some of the reported KG completion results are significantly lower than what they potentially could be.
Evaluation of the machine learning tasks that use link information from the knowledge graphs and neural embeddings is explored in~\cite{ristoski_2016,cochez_2017}.

As a general remark, in most of the works, link prediction is evaluated with the mean rank metric and its variants from the information retrieval community  (e.g., mean
reciprocal rank, top $k$ results, mean average precision), which we believe is not the most suitable metric for link prediction. As we pointed out in
Section~\ref{sec:intro-lp}, the probability to have a new link $(e_i, r_i, e_j)$ is different from asking if the entity $e_j$ is part of the set of the most probable
entities, among all existing entities in the knowledge graph, to be connected to $e_i$ with the relation $r_i$.

We believe that the evaluation of the quality of the embeddings for the link prediction task has received much less attention than the methodologies for training the
embeddings in the literature. While most of the works do perform extensive evaluation of their embedding approaches, the exact steps and implications of negative sample
generation, random train and test data splits, amount of information involved in the unsupervised learning, are either not very well detailed for an easy and fair
reproducibility of the results, or are presented as a secondary remark.

\subsubsection{Negative example generation for link prediction} 
In general, the link prediction problem for knowledge graphs is different from other classification problems where positive and negative examples are well defined.
Obtaining a representative test set with a prototypical distribution is often not trivial~\cite{garciagasulla_2016}, and usually what is done is that we randomly remove
some links which we then use as our test positives. Moreover, during the generation of negative links both for train and test sets, we impose that no negative link
appears as a training positive or test positive. We therefore implicitly leak information about the test positives when we generate train negatives. In other words,
during the generation of negative links $(\bar{e_i}, r_i, \bar{e_j}) \not \in KG)$ we should account to the possibility that this link might actually turn out to be true,
and our binary classifiers should be robust and generalize well to these realistic situations. As our future work we would like to study further the implications of the
negative example generation.

\section{Conclusions}
In this paper we focus on link prediction for knowledge graphs, treated as a binary classification problem on entity embeddings. In this work we provide our first results
of the evaluation of different strategies for training neural embeddings of entities on the WN11 knowledge graph. These early findings lead us to suggest that: i) if the
number of multi-relational connectivities of nodes is low compared to the total number of connections, then the graph can be \emph{flattened} and treated as an unlabelled
graph (i.e., no two nodes are connected with more than one link), provided that we disambiguate the links with separate binary classifiers for each relation $r_i$; ii)
training embeddings once globally and using them in binary classifiers for each relation $r_i$ gives comparable classification error (F-measure $\approx 0.8$ averaged
over all relations on WN11) as embeddings trained locally for each relation separately. The global approach to training embeddings is more scalable as it requires only
one neural model to represent all entities, as opposed to having as many models as there are many relations in the knowledge graph. The confidence in our results is, of
course, proportionate to the amount of information (percentage of all available links) that we include in the unsupervised training. Depending on the incoming and
outgoing degrees of the entities in the graph, the global approach may fail to embed many entities. Thus, the global approach is less robust to limited availability of
information then the local approach. Finally, we make our code for the evaluation pipeline for link prediction tasks open source~\cite{github_repo}, and hope that it will
trigger a standardized benchmark for the evaluation of the knowledge graph embeddings.

%
%

\end{document}